\documentclass[journal]{IEEEtran}
\usepackage{amsmath,amsfonts}
\usepackage{algorithmic}
\usepackage{algorithm}
\usepackage{array}
\usepackage[caption=false,font=normalsize,labelfont=sf,textfont=sf]{subfig}
\usepackage{textcomp}
\usepackage{stfloats}
\usepackage{url}
\usepackage{verbatim}
\usepackage{graphicx}
\usepackage{cite}
\usepackage{booktabs}
\usepackage{multirow}
\usepackage[table]{xcolor}
\usepackage{soul} % 导入 soul 包
\usepackage{color, xcolor}
\usepackage{hyperref}
\usepackage{hyperref}\hypersetup{hidelinks,colorlinks=true,allcolors=black,pdfstartview=Fit,breaklinks=true}
\begin{document}

\title{NeuV-SLAM: Fast Neural Multiresolution Voxel Optimization for RGBD Dense SLAM}

\author{Wenzhi Guo, Bing Wang$^{\ast}$, Lijun Chen$^{\ast}$
        % <-this % stops a space
\thanks{W. Guo is with the Department of Computer Science and Technology, Nanjing University, Nanjing, China, and the Department of Aeronautical and Aviation Engineering, The Hong Kong Polytechnic University, Hong Kong, China}% <-this % stops a space
\thanks{B. Wang is with the Department of Aeronautical and Aviation Engineering, The Hong Kong Polytechnic University, Hong Kong, China}
\thanks{L. Chen is with the Department of Computer Science and Technology, Nanjing University, Nanjing, China}

}

\markboth{Journal of \LaTeX\ Class Files,~Vol.~14, No.~8, Jan~2024}%
{Shell \MakeLowercase{\textit{et al.}}: A Sample Article Using IEEEtran.cls for IEEE Journals}


\maketitle

\begin{abstract}
We introduce NeuV-SLAM, a novel dense simultaneous localization and mapping pipeline based on neural multiresolution voxels, characterized by ultra-fast convergence and incremental expansion capabilities. This pipeline utilizes RGBD images as input to construct multiresolution neural voxels, achieving rapid convergence while maintaining robust incremental scene reconstruction and camera tracking. Central to our methodology is to propose a novel implicit representation, termed \textit{VDF} that combines the implementation of neural signed distance field (SDF) voxels with an SDF activation strategy. This approach entails the direct optimization of color features and SDF values anchored within the voxels, substantially enhancing the rate of scene convergence. To ensure the acquisition of clear edge delineation, SDF activation is designed, which maintains exemplary scene representation fidelity even under constraints of voxel resolution. Furthermore, in pursuit of advancing rapid incremental expansion with low computational overhead, we developed \textit{hashMV}, a novel hash-based multiresolution voxel management structure. This architecture is complemented by a strategically designed voxel generation technique that synergizes with a two-dimensional scene prior. Our empirical evaluations, conducted on the Replica and ScanNet Datasets, substantiate NeuV-SLAM’s exceptional efficacy in terms of convergence speed, tracking accuracy, scene reconstruction, and rendering quality. Our code will be available at \href{https://github.com/DARYL-GWZ/NeuV-SLAM/tree/main}{https://github.com/DARYL-GWZ/NeuV-SLAM/tree/main}. 

\end{abstract}


\section{Introduction}
\IEEEPARstart{D}{ense} Simultaneous Localization and Mapping (SLAM) stands at the forefront of computational perception, offering the dual capabilities of estimating the pose of a camera and meticulously creating a detailed topographical representation of the surrounding environment. This technique has been smoothly integrated into many different applications, ranging from the automation of vehicular navigation to the nuanced intricacies of augmented reality experiences \cite{huang2019apolloscape,manuelli2019kpam,marion2018label,hodan2018bop}.\par
Conventional SLAM methodologies, predominantly based on explicit spatial representations such as point clouds\cite{mur2015orb,mur2017orb,campos2021orb}, voxel grids\cite{kahler2015very,newcombe2011kinectfusion}, and surfel \cite{wang2019real,stuckler2014multi}, often struggle with capturing detailed color and luminance, leading to inconsistent or sparsely populated maps. This undermines the precision and reliability of the mapping process, making it crucial to overcome these challenges. Neural Radiance Fields (NeRF) \cite{mildenhall2021nerf}, utilizing deep neural networks, present a compelling solution by providing a compact yet powerful representation capable of rendering photorealistic scenes with minimal storage requirements.  Initial explorations into NeRF-based SLAM methodologies\cite{sucar2021imap}, \cite{zhu2022nice} have demonstrated their potential in providing detailed and continuous scene representations for enhancing dense RGBD SLAM frameworks.\par
\begin{figure}[!t] 
\centering %
\includegraphics[width=0.5\textwidth]{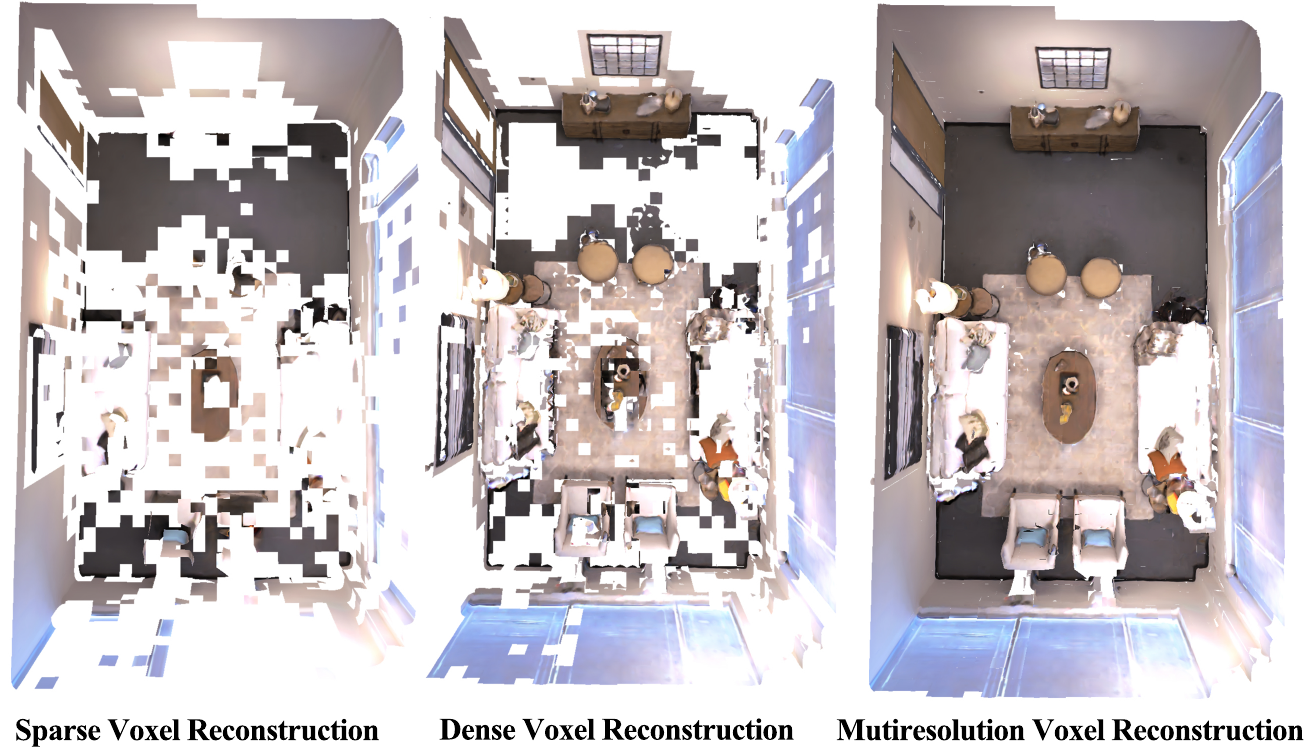} 
\caption{We propose NeuV-SLAM, a SLAM system that incrementally reconstructs scenes from sequential RGBD frames. NeuV-SLAM reconstructs the scene separately based on dense voxels and sparse voxels. } 
\vspace{-0.3cm}
\label{top} 
\end{figure}
The majority of existing typical implicit SLAM pipelines, as exemplified by \cite{zhu2023nicer}, assume the spatial partitioning of scenes into grids \cite{yan2023gs} and depend on multi-layered grids and complex neural networks for scene representation, which limits their capability for scene expansion. Additionally, the complexity of these methods, coupled with their reliance on prior information or pre-trained decoders, hampers their ability to efficiently represent scenes. This significantly restricts their applicability in larger-scale and more complex scenes. Therefore, a question remains unanswered: \textbf{How can one scalably and efficiently learn the scenes through sequential RGBD frames}? \par
Implementing an incrementally scalable SLAM system in implicit environments presents significant difficulties. As RGBD image sequences are inputted, signifying the increment expansion of the scene, the SLAM system must be capable of real-time integration of new scene elements into the model while ensuring overall scene consistency and low computational resources. This necessitates an efficient management structure within the system that supports scene expansion. Recent research, as indicated in references \cite{yang2022vox,mao2023ngel}, has experimented with the use of voxel representations for scene depiction and dynamic management through octree structures to achieve real-time expansion and reconstruction of unknown scenes. However, due to the hierarchical segmentation inherent in octree structures, they face challenges of high structural complexity and excessive data volume, which have become primary obstacles in scene expansion.\par
Moreover, there remain significant challenges in achieving efficient learning of scenes. The primary issue lies in the limited capacity of neural networks. As the scene expands, the network often faces the problem of forgetting previously learned scene elements while learning new ones. This leads to difficulty in forming a consistent representation of the entire scene. Additionally, existing implicit SLAM systems often employ volume density methods to capture scene surfaces, but these methods are not always direct and efficient. Also, the use of a single-size voxel grid struggles to balance between rapid and accurate scene capture and the consumption of computational resources. For instance, the Vox-Fusion\cite{yang2022vox} method attempts to store neural features directly in voxel vertices to accelerate the optimization process of the scene, but this approach still relies on larger-scale neural networks for accurate scene representation. Additionally, Point-SLAM \cite{sandstrom2023point} achieves precise capture of scenes by using point clouds. However, this method struggles with fast processing when optimizing a large number of neural points and incurs substantial computational resource consumption. \par
\begin{figure*}[!t]
\centering
\includegraphics[width=7.1in]{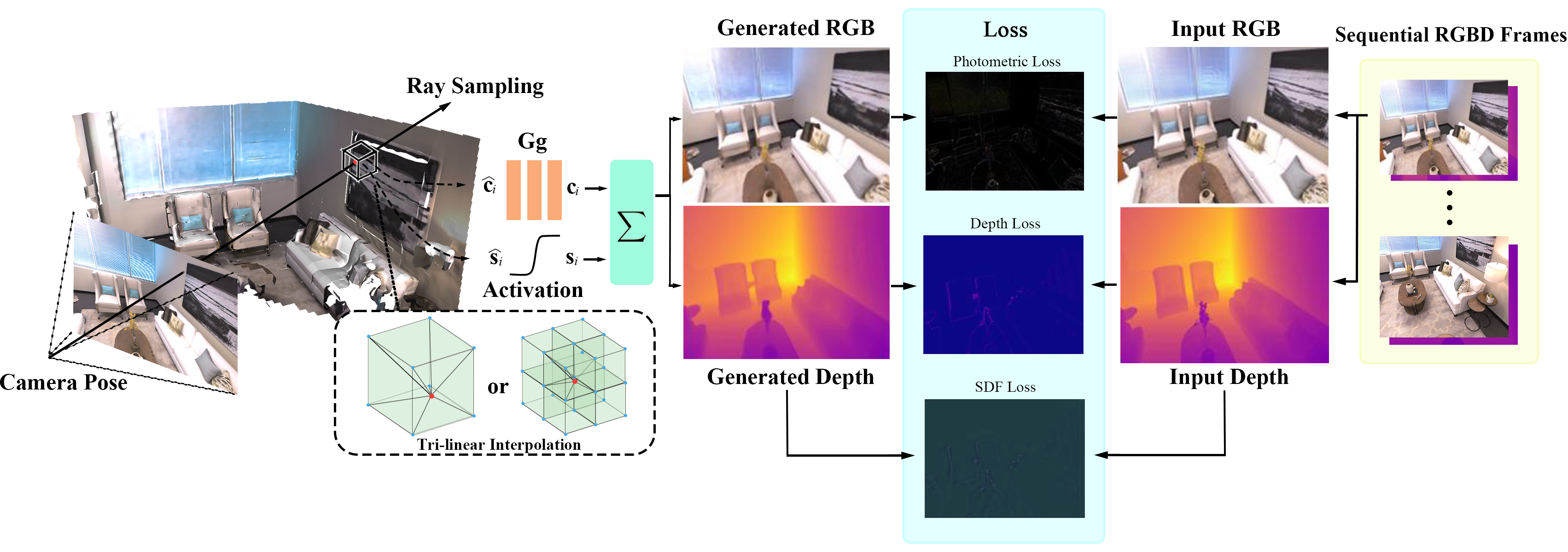}
\caption{\textbf{Overview of NeuV-SLAM}. NeuV-SLAM takes RGBD images as input and directly anchors color features and SDF  values in multiresolution voxels to estimate camera pose and learn scene representation. From left to right, during the mapping stage, SDF values obtained directly through activated trilinear interpolation efficiently learn scene geometry, and scene color information is learned through interpolated neural features. The depth and color values are rendered through volumetric rendering, minimizing color, depth, and SDF losses to optimize the network \textit{$G_g$}. From right to left, during the tracking stage, \textit{$G_g$}  parameters are fixed, and the camera pose is updated through backward propagation. Incremental expansion of the scene is achieved through the hashMV structure. The tracking and mapping stages alternate until the entire SLAM process is completed, with the multiresolution voxels converging to a finite set.}
\label{main}
\end{figure*}

To tackle all these challenges, we present NeuV-SLAM, a dense SLAM system based on neural multiresolution voxels, to achieve efficient incremental scene expansion, as shown in Figures \ref{top} and \ref{main}.  It consists of two major components: 1) We design an efficient hash-based multiresolution voxel management structure, termed \textbf{\textit{hashMV}}, which robustly supports rapid dynamic scene expansion. Leveraging the innovative structure, our system achieves efficient scene reconstruction and incremental expansion in unknown environments while maintaining a small memory footprint. 2) To achieve efficient scene convergence, we propose an innovative implicit scene representation method, named \textbf{\textit{VDF}}, that anchors color features and SDF values directly within multiresolution voxels and employs an SDF activation strategy to enhance the capability of capturing finer scene details. Consequently, we are able to utilize a convergence-friendly lightweight decoder to learn scene colors by minimizing depth, color, and SDF losses, thus alternately optimizing the decoder and camera poses during the tracking and mapping phases. Finally, we comprehensively evaluate our method and the experiments demonstrate its competitiveness in terms of convergence speed, reconstruction quality, tracking accuracy, and rendering performance.

To summarize, our main contributions are as follows:
\begin{itemize}
\item{We present NeuV-SLAM, an innovative RGBD dense SLAM based on neural multiresolution voxels, designed to efficiently accomplish incremental scene reconstruction and camera pose estimation in unknown environments.}
\item{We develop a novel multiresolution voxel management architecture, named \textbf{\textit{hashMV}}, which is based on 2D information and hash structures. This facilitates efficient and scalable management of scenes.}
\item{We propose a novel implicit scene representation, named \textbf{\textit{VDF}}, incorporating neural SDF voxels and SDF activation. This method involves directly anchoring SDF within multiresolution voxels and employing the activation function, enhancing the convergence efficiency and augmenting the ability to accurately fit detailed scenes. }
\item{We conduct evaluations of our method using the Replica and ScanNet Datasets. Our approach exhibits strong competitiveness in convergence speed, reconstruction, tracking, and rendering capabilities.}
\end{itemize}
\section{Related Work}
\subsection{Dense SLAM}
 Contemporary advancements in dense SLAM systems have bifurcated predominantly into two distinct paradigms: conventional SLAM methodologies and those predicated on learning-based approaches. The former, extensively elucidated in seminal works \cite{klein2008improving,mur2017orb,davison2007monoslam,klein2007parallel,engel2017direct,newcombe2011dtam,engel2014lsd,qin2018vins,xu2021fast,xu2022fast,zhou2023backpack,zhou2023asl}, predominantly harness intricate algorithms to facilitate real-time tridimensional reconstruction. Conversely, the latter paradigm, as expounded in \cite{yokozuka2019vitamin,saputra2020deeptio,li2020deepslam,iegawa2023loop,zhou2023lightweight,memon2020loop,mukherjee2019detection}, integrate machine learning techniques to augment the efficacy of the scene reconstruction process.

In the evolution of traditional SLAM frameworks, KinectFusion \cite{izadi2011kinectfusion} simplifies indoor 3D reconstruction but is limited to small, indoor environments. ElasticFusion \cite{whelan2015elasticfusion} and LSD-SLAM \cite{engel2014lsd} offer better adaptability for large-scale mapping, with the former excelling in unmarked areas and the latter on lower-end hardware. However, they may struggle with dynamic objects and lighting changes. DVO SLAM \cite{yan2017dense} and DTAM \cite{newcombe2011dtam} are ideal for accurate depth estimation using photometric alignment but require significant computational power. ORB-SLAM3 \cite{campos2021orb} is efficient and precise in feature-rich areas but less so in sparse or fast-moving scenes. InfiniTAM \cite{prisacariu2017infinitam} and Voxblox \cite{oleynikova2017voxblox} support diverse, large-scale environments but need careful calibration. OpenLORIS-Scene \cite{shi2020we} focuses on dense mapping in dynamic settings, best used within a comprehensive and complex system.

In recent years, there has been a series of research studies proposing the replacement of certain components of traditional SLAM systems with learning-based modules. For instance, convolutional neural networks (CNNs) have been trained to perform stereo matching by learning similarity metrics from small image patches, thereby extracting depth information from image pairs with enhanced accuracy and efficiency \cite{zbontar2016stereo}. Similarly, novel deep network architectures have been developed for end-to-end feature processing, including detection, orientation estimation, and feature description, surpassing traditional methods like SIFT \cite{wu2013comparative} in density and availability \cite{yi2016lift}. Additionally, deep learning has been leveraged for odometry estimation, combining visual and inertial data to estimate motion without the need for laborious manual synchronization or calibration between cameras and IMUs \cite{clark2017vinet}. In summary, these systems demonstrate the diversity and flexibility of dense SLAM techniques based on traditional explicit representations. However, explicit representations fundamentally limit the applicability of the above-mentioned systems.
\vspace*{-2mm}
\subsection{Neural Implicit Representation}
The quest to accurately represent three-dimensional scenes in computer vision constitutes a significant scholarly challenge. This field has seen substantial contributions, as evidenced in advanced works \cite{oleynikova2017voxblox,reijgwart2019voxgraph,tancik2020fourier}. To overcome the discretization issues of explicit representations, recent advancements have primarily focused on neural network-based implicit scene representation methods. These implicit representations can typically be categorized into, for example, signed distance fields \cite{ park2019deepsdf}, occupancy fields \cite{ mescheder2019occupancy}, unsigned distance fields \cite{ chibane2020neural}, and radiance fields \cite{mildenhall2021nerf}, which have found widespread applications in several domains. These include 3D reconstruction, as highlighted in \cite{sitzmann2020implicit,oechsle2021unisurf,chen2021mvsnerf,peng2020convolutional}, novel view synthesis, as explored in \cite{zhang2020nerf++,martin2021nerf}, and in 3D generative models, as indicated in \cite{niemeyer2021giraffe,schwarz2020graf,chan2021pi}. The advent of Neural Radiance Fields (NeRF) \cite{mildenhall2021nerf} has particularly catalyzed advancements in this area. The efficacy of neural implicit representation in rendering extensive environments with precision is well-established, as demonstrated in studies \cite{rematas2022urban, turki2022mega,tancik2022block}.

Despite its innovations, neural implicit representation faces significant challenges, primarily its protracted training and rendering times. In response, the academic community has developed various strategies to enhance efficiency, including advanced precomputation mechanisms \cite{garbin2021fastnerf,yu2021plenoctrees,reiser2021kilonerf}. Efforts have also been made to optimize NeRF’s multi-layer perceptrons for better processing of sparse scene features, albeit often at the expense of increased memory requirements, as discussed in \cite{deng2020jaxnerf,liu2020neural,song2019autoint}. 
Nevertheless, addressing the efficiency issues within NeRF-SLAM systems remains a significant challenge.
% In this paper, we leverage the success of implicit representations and introduce a novel implicit representation approach to enhance the efficiency of the system.
\vspace*{-2mm}
\subsection{NeRF based SLAM System}
In the domain of NeRF-based SLAM, a prominent approach involves using Neural Radiance Fields for simultaneous camera pose estimation and environmental reconstruction. A significant contribution in this area is BARF \cite{lin2021barf}, which integrates the optimization of NeRF model parameters and camera poses. This method employs adaptive masked position encoding, enabling a more refined coarse-to-fine registration process. Another groundbreaking work is iMAP \cite{sucar2021imap}, which stands out as the first NeRF-based online dense SLAM model. This model excels in concurrently optimizing camera poses and implicit scene representations, facilitating continuous online learning. Building on iMAP, NICE-SLAM \cite{zhu2022nice} introduces key improvements in keyframe selection and NeRF structure, leading to enhanced scene details, quicker tracking and mapping, and more accurate pose estimation. An extension of NICE-SLAM is NICER-SLAM, a remarkable RGB-only SLAM system. Vox-Fusion \cite{yang2022vox} integrates neural implicit representations with traditional volume fusion methods, utilizing voxel-based neural implicit surface representations to encode and optimize scenes within each voxel. Point-SLAM \cite{sandstrom2023point}  using monocular RGBD input, anchored in a neural scene representation within a dynamically generated point cloud. This method enables both tracking and mapping with a point-based neural scene representation, optimizing for RGBD-based re-rendering loss. It dynamically adapts anchor point density based on the information density of the input, enhancing runtime and memory efficiency. In addition, The integration of traditional SLAM systems with NeRF-based methods highlights the significant potential of such combined approaches in the field of dense slam \cite{rosinol2022nerf,chung2023orbeez}. \par
We conducted a thorough analysis of traditional SLAM technologies, elucidating the limitations of conventional explicit representations in terms of scene reconstruction accuracy. To push the boundaries of these traditional frameworks, we integrate them with cutting-edge learning methodologies, aiming to enhance the adaptability and overall performance of SLAM systems within diverse and complex environments. The current study specifically focuses on harnessing the potential of neural implicit representations for map representation within dense SLAM tasks. We posit that by integrating advanced scene reconstruction techniques inherent in neural implicit representations, we can achieve more precise, efficient, and lifelike map generation, thereby significantly elevating the operational efficiency of autonomous systems navigating complex spaces. Building on these technological advancements, our research introduces an innovative implicit  SLAM framework based on neural multi-resolution voxels. This carefully crafted framework aims to significantly improve operational efficiency and scalability, making it highly effective in diverse and complex environments.
\section{NeuV-SLAM}
\subsection{Multiresolution Voxel Generation and Management}
\begin{figure*}[!t] 
\centering %
\includegraphics[width=0.72\textwidth]{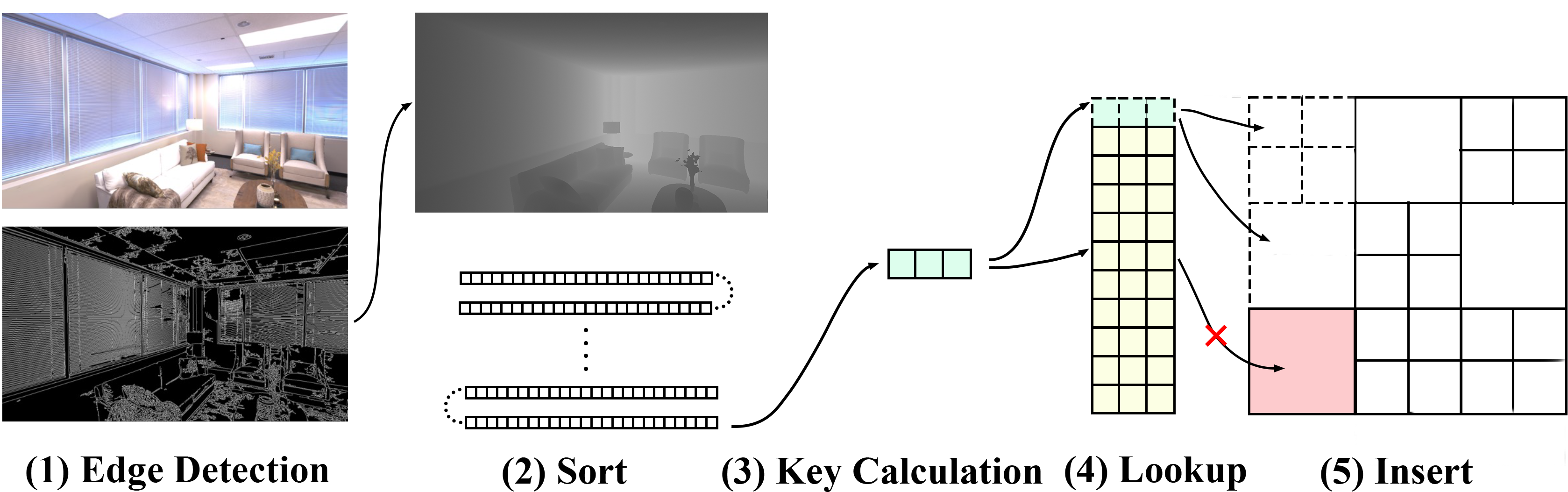} 
\caption{\textbf{The process of multiresolution voxel generation.} \textbf{Edge Detection}: This step entails identifying boundaries within an image by pinpointing areas of significant brightness discontinuities. \textbf{Sort}: The detected edge points are then prioritized and resequenced, with a focus on these points for subsequent processing. \textbf{Key Calculation}: Utilizing the positional data of each point, a unique identifier or 'key' is computed. \textbf{Lookup}: This key is used to search within a hash table. Absent keys prompt new key generation.
\textbf{Insert}: Voxel creation is guided by these keys. Edge points lead to denser voxel formation, while non-edge points result in sparser voxels. Existing occupied spaces result in key disposal.
} 
\label{add} 
\end{figure*}
\subsubsection{Generation}
Contrary to the \cite{zhu2022nice,yang2022vox} approach, which employs a single-resolution voxel, we utilize multiresolution voxels based on scene details to represent the scene. This ensures efficient and accurate scene fitting with minimal computational cost. The efficacy of multi-resolution voxels is demonstrated in the ablation experiment F 2). In the mapping phase, subsequent to the successful tracking of a new frame, we execute an inverse projection of its depth estimation into a three-dimensional spatial domain. This is achieved by utilizing the pose of the current frame to translocate these tridimensional points into the global coordinate framework. Subsequently, these spatial points are annotated with edge attributes, derived from the edge information intrinsic to the current frame. The operational principle for voxel generation adheres to a 'first-come, first-served' protocol in relation to spatial point processing. Emphasizing the edge scene, a prioritization algorithm is employed for sorting these tridimensional points, preferentially processing those endowed with edge attributes. This methodology facilitates the generation of densely populated voxels, thereby mitigating the prospective overshadowing of dense voxels by their sparser counterparts. Following this, we compute a unique identifier (\textit{Key}) for each point in the three-dimensional space, employing the subsequent methodology:
\begin{equation}
key=floor(\frac{{p}_{(x, y ,z)}}{v}).
\end{equation}
The \textit{Key} serves as a hash index to identify the positional information of voxel, while $p_{(x,y,z)}$ represents the positional information of space points, and \textit{v} signifies the edge length of voxels. This process involves querying the hash table: if the queried \textit{Key} is already present in the table, it indicates a duplication of the corresponding spatial point, leading to its exclusion; conversely, if the \textit{Key} is absent, a new voxel is allocated to the spatial point based on edge information. Within this framework, edge spatial points result in the generation of dense voxels, whereas regular spatial points produce sparse voxels. Notably, the resolution of dense voxels is twice that of sparse voxels. This design is implemented to optimize the balance between spatial resolution and processing efficiency.
\subsubsection{Management}
Diverging from the approach in existing SLAM systems, which predominantly utilize a static single resolution and employ an axis-aligned voxel division approach for the entire scene \cite{zhu2022nice}, our method implements an innovative incremental scene expansion strategy, along with a multiresolution voxel representation to preserve a greater level of detail information. Drawing inspiration from the design of traditional SLAM systems, we manage multiresolution voxels using a hash table, enabling incremental expansion in unexplored areas. We have named this structure, \textbf{\textit{hashMV}}; it facilitates the rapid dynamic addition of multiresolution voxels and the effective retrieval of adjacent voxels, significantly enhancing the efficiency of voxel generation and retrieval.\par
In the process of adding multiresolution voxels, we assign a unique number to each voxel vertex, while ensuring that these numbers are shared with adjacent voxels. Owing to the disparate resolutions of voxels, acquiring the numbers of neighboring voxels is typically complex. To resolve this challenge, we have established a correspondence between spatial positions and voxel numbers. Initially, shared numbers are determined based on the relative positions of the new voxel to its adjacent voxels, and these are assigned to the appropriate spatial locations of the new voxel. Subsequently, the remaining positions are numbered in a predefined sequence. This strategy of number sharing not only saves memory resources but also significantly expedites the convergence speed of neural voxels.
\vspace*{-7mm}
\subsection{Fast Neural Multiresolution Voxel Optimization}
\subsubsection{Multiresolution Points Sampling}
In the pursuit of optimizing point sampling efficacy within our framework, our methodology entails the deployment of rapid ray tracing during the processing of each sampling ray. This is executed to ascertain the intersectionality of the ray with both sparse and dense voxel structures, thereby determining the occurrence of any interaction. In instances where a ray fails to intersect with the voxel ensemble, it is deduced that the corresponding pixel lacks contributory significance to the resultant rendering output, prompting its exclusion from the rendering pipeline. For rays that do exhibit intersections with voxels, uniform sampling is conducted along the trajectory between the ray-voxel intersection points, utilizing a predetermined step length denoted by \textit{S}. This procedure yields sets of space points corresponding to the sparse and dense voxel domains. Finally, we merge the sampling points located on the same ray in both sparse and dense voxels and sort them by depth to facilitate subsequent volume rendering operations.
\subsubsection{Neural Multiresolution Voxel Representation}
Diverging from Vox-Fusion \cite{yang2022vox} and DVGO \cite{sun2022direct} methodologies, our voxel grid representation employs trilinear interpolation for simultaneous modeling of SDF and color features within voxel cells, which enhances precision in querying any space position, significantly boosting scene convergence efficiency:
\begin{equation}
\text { interp }(\boldsymbol{x},\boldsymbol{{V}^{(D)}},\boldsymbol{{V}^{(S)}}):\boldsymbol{x} \in \mathbb{R}^{3},\boldsymbol{{V}^{(D)}},\boldsymbol{{V}^{(S)}} \in \mathbb{R}^{A \times N}.
\end{equation}
\par
\textit{x} represents the 3D positional coordinates of the space point, \textit{V} denotes the voxel grid, \textit{A} is the dimension of the SDF or color feature, \textit{N} is the number of dense or sparse voxels.
\subsubsection{SDF Activation Strategy}
The SDF voxel $V^{SDF}$ is utilized for storing SDF values in volumetric rendering. Within this framework, we use  \textit{$\hat{s}$} to represent the original voxel SDF value before processing through the SDF activation. To enhance the representational capability of the SDF voxel grid while maintaining the intrinsic properties of SDF, we employ the hyperbolic tangent function (tanh) as the activation function to process SDF values \textit{$\hat{s}$}, as follows:
\begin{equation}
s=\tanh \hat{s}=\frac{e^{\hat{s}}-e^{-\hat{s}}}{e^{\hat{s}}+e^{-\hat{s}}}.
\end{equation}
The application of the hyperbolic tangent function (Tanh) enables the effective exploration of SDF values that are less than zero, concurrently facilitating the non-linearization of SDF values with the same sign. This methodology optimizes sensitivity to minute variations and enhances the model's precision in handling regions proximate to surfaces.

The interpolated values of the voxel SDF are subjected to a sequential processing regimen, involving the hyperbolic tangent function (tanh) and an interpolation function (interp). Taking inspiration from the DVGO \cite{sun2022direct} and its post-activation strategy, the interpolation of voxel SDF values is sequentially processed using the interp and tanh functions for volume rendering. This method enhances the ability to produce well-defined, sharp surfaces, significantly improving the voxel grid's capacity to accurately capture and represent intricate geometric details. The formula of $s^{(post)}$ is as follows:
\begin{equation}
\begin{aligned}
s^{\text {(post) }} & =\text { tanh }\left(\text { interp }\left(\boldsymbol{x}, \boldsymbol{V}^{\text {(SDF) }}\right)\right).
\end{aligned}
\end{equation}
\subsubsection{Volume Render}
The model can be conceptually abstracted as an advanced neural network module. This module synthesizes the data structures of \textit{m} sparse voxels and \textit{n} dense voxels, combined with \textit{j} view-dependent sampled rays. These rays further sample \textit{i} spatial points to acquire their positional information. Based on this structure, the module is capable of accurately regressing and analyzing the RGB color values and SDF values for each ray at specific positions, as follows:
\begin{footnotesize}
\begin{equation}
\left({D}_{j}, {C}_{j}\right)=NeuV-SLAM\left(x_1, \boldsymbol{V}_1^S, \boldsymbol{V}_1^D, \ldots,x_i, \boldsymbol{V}_m^S, \boldsymbol{V}_n^D\right).
\end{equation}
\end{footnotesize}
In contrast to existing methods like Point-SLAM \cite{sandstrom2023point} and Vox-Fusion \cite{yang2022vox}, which rely on neural networks to concurrently regress both identity and RGB values during their processing phases, this study introduces a more rapid and streamlined implicit representation approach, named \textbf{\textit{VDF}}. Our focus lies in the precise regression of RGB values for sampled points using a lightweight network designed for fast convergence, while directly optimizing the scene geometry through the neural SDF voxel grid. 
\begin{figure}[!t] 
\centering %
\includegraphics[width=0.32\textwidth]{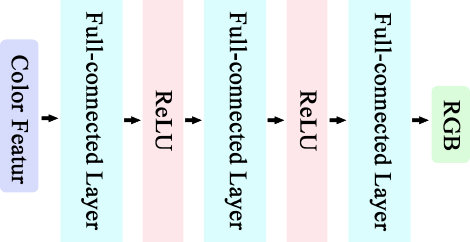} 
\caption{The lightweight network architecture of $G_g$.} 
\label{net} 
\end{figure}
\par
Specifically, we perform trilinear interpolation and activation strategy on the SDF dense voxel grid $V^{(D)(SDF)}$ and SDF sparse voxel grid $V^{(S)(SDF)}$ based on the positional coordinates of sampled points. This process allows us to approximate and reconstruct the complex geometric structures of the scene with accuracy and continuity. Additionally, to model color attributes independent of viewing angles, we introduce other voxel grids, $V^{(D)(RGB)}$ and $V^{(S)(RGB)}$. These grids are specifically designed to store and process color information, enabling precise modeling of scene color attributes. Consequently, we efficiently obtain SDF and RGB values for any spatial point in the scene, as follows:

\begin{small}
\begin{equation}
\begin{aligned}
& \boldsymbol{\hat{c_i}}=\operatorname{interp}\left(\boldsymbol{x_i}, \boldsymbol{V}^{(\mathrm{S})(\mathrm{RGB})}\right)\cup\operatorname{interp}\left(\boldsymbol{x_i}, \boldsymbol{V}^{(\mathrm{D})(\mathrm{RGB})}\right),\\
& \boldsymbol{\hat{s_i}}=\operatorname{interp}\left(\boldsymbol{x_i}, \boldsymbol{V}^{(\mathrm{S})(\mathrm{SDF})}\right)\cup\operatorname{interp}\left(\boldsymbol{x_i}, \boldsymbol{V}^{(\mathrm{D})(\mathrm{SDF})}\right), \\
& s_i  =\text { tanh }\hat{s_i}.
\end{aligned}
\end{equation}
\end{small}

Subsequently, we employ a decoder, denoted as \textit{$G$}, characterized by a set of tunable parameters represented as \textit{g}, for the purpose of regressing RGB values. The architectural blueprint of this network is elucidated in Figure \ref{net}. During this procedural step, we utilize the color features, denoted as \textit{$\hat{c}_i$}, of the sampled points as input, facilitating the acquisition of the respective color values associated with each sampled point. The equation of $c_i$ is as follows:
\begin{equation}
c_i=G_g\left(\hat{c}_i\right).
\end{equation}

Following this, we perform a holistic processing of the sampled points situated along the same ray, prioritizing them based on their depth, to ensure orderly processing:
\begin{equation}
r_j=\operatorname{sort}\left(r_j^D \cup r_j^S\right).
\end{equation}

Then, we adopt a volumetric rendering technique similar to the Vox-Fusion \cite{yang2022vox} to compute and render depth \textit{$D$} and color \textit{$C$} attributes along the entire ray path. This process entails a meticulous examination of the gradient fluctuations within the SDF, facilitating the accurate identification of sampled points situated in close proximity to the neighboring regions of the scene's geometric surfaces. Additionally, we employ an effective strategy to exclude points that lie along the path between the scene surfaces and the camera sensor, thereby optimizing the efficiency of point sampling selection. The mathematical expression for this process is as follows:
\begin{equation}
\begin{aligned}
o_i & =sigm\left(\frac{s_i}{st}\right) \cdot sigm\left(-\frac{s_i}{st}\right), \\
D_j & =\frac{\sum_{i=0}^{N-1} o_i \cdot d_i}{\sum_{i=0}^{N-1} o_i},\\
C_j & =\frac{\sum_{i=0}^{N-1} o_i \cdot \mathbf{c}_i}{\sum_{i=0}^{N-1} o_i}.
\end{aligned}
\end{equation}

$sigm(.)$ represents the sigmoid function, where \textit{$d_i$} stands for the depth of the sampled point, and \textit{$st$} denotes the predefined SDF truncation distance, \textit{$s_i$} is the predicted SDF value.
\vspace*{-3mm}
\subsection{Tracking and Mapping}
\subsubsection{Tracking}
In the tracking phase, we employed a replica of the implicit network and neural voxels created during the mapping phase, with their parameters fixed. To predict the current frame's camera pose, we utilized a basic zero-motion model, which estimates the six degrees of freedom (6-DoF) camera pose e $\in$ SE(3) starting from the known pose of the most recent frame. Subsequently, we refined the optimization of e$\in$SE(3) through a series of precise computational steps—including sampling of the scene, efficient volumetric rendering, and rigorous minimization of the rendering loss.
\subsubsection{Mapping}
In the mapping phase, our method involves the random selection of \textit{$N$} rays for analysis within the current RGBD frame. The essence of this process lies in the precise minimization of color, depth, and SDF losses to guide the optimization of the implicit network and neural voxel features. To be specific, the color and depth losses are computed by contrasting the colors and depths acquired through volumetric rendering with the actual values observed along the rays in the real-world scene. Simultaneously, the SDF loss quantifies the deviation between the rendered depth and the actual depth, taking into account the SDF values. The mathematical expression for this loss function is as follows:

\begin{footnotesize}
\begin{equation}
\begin{aligned}
L=&\frac{1}{|R|} \sum_{i=0}^{|R|}\left\|\mathbf{C}_i-\mathbf{C}_i^{g}\right\|+
\frac{1}{|R|} \sum_{i=0}^{|R|}\left\|D_i-D_i^{g}\right\|\\
&+ \frac{1}{|R|} \sum_{R \in p} \frac{1}{S_p^{st}} \sum_{s \in S_p^{t r}}\left(D_s-D_s^{g}\right)^2.
\end{aligned}
\end{equation}
\end{footnotesize}

This method goes beyond the consideration of color and depth consistency and, critically, provides a precise evaluation of the congruence between the geometric information and the real-world scene. This rigorous approach guarantees a high level of accuracy and meticulous preservation of fine details in the scene reconstruction process.

In addition, we employ a mesh grid for scene reconstruction. Initially, a uniform sampling strategy is applied within both the sparse voxel and dense voxel spaces to generate a set of sampled points. Subsequently, while keeping the parameters of the implicit network and neural voxels fixed, we determine the SDF values through the application of trilinear interpolation and activation strategy. Following this, the Marching Cubes algorithm \cite{lorensen1998marching} is utilized to process these SDF values, thereby ascertaining the positions of vertices and the corresponding facets, which are used to construct the mesh representation. Finally, the vertices and facets generated within the sparse voxel and dense voxel spaces are combined to form the geometric structure of the scene.
\section{Experiment}
\subsection{Experimental Setup}
\subsubsection{Datasets}

Our research was carried out using the Replica and ScanNet Datasets. The Replica Dataset is renowned for its 18 photorealistic indoor scenes, featuring a variety of environments such as offices, living rooms, bedrooms, and kitchens. In line with the experimental setup used in Vox-Fusion, we utilized the same 8 scenes from the Replica Dataset for our testing and performance assessment. The ScanNet Dataset, on the other hand, offers a comprehensive array of RGBD data sourced from actual indoor settings. For our evaluation, we selected a range of scenes from this dataset, carefully chosen to cover a broad spectrum of challenges typically found in real-world indoor environments.

\subsubsection{Baseline}
For our comparative analysis, we employed iMAP \cite{sucar2021imap}, Vox-Fusion \cite{yang2022vox}, NICE-SLAM \cite{zhu2022nice}, and DI-Fusion \cite{huang2021di} as our baseline methods. These systems have been recognized for their exceptional capabilities in scene reconstruction and camera pose estimation, making them ideal references for evaluating the effectiveness of our approach.
\subsubsection{Metrics}
In our systematic evaluation process, we employed a set of metrics to thoroughly gauge the quality of the system. These included mesh accuracy and completion metrics to evaluate reconstruction quality. Mesh accuracy (Acc.) is defined as the non-directional chamfer distance from the reconstructed mesh to the ground truth. Completion (Comp.) is also defined as the distance in the opposite direction. To measure the accuracy of camera tracking, we used the absolute trajectory error (ATE) metric. For assessing the quality of rendering, we relied on metrics such as the Peak Signal-to-Noise Ratio (PSNR), Multi-Scale Structural Similarity Index (SSIM), and Learned Perceptual Image Patch Similarity (LPIPS). Together, these metrics form a comprehensive framework for assessing the accuracy and quality of our scene reconstruction and camera pose estimation techniques, providing a basis for effective comparison and in-depth analysis of our method's performance. In addition, The codes for tracking and mapping evaluation are both from Vox-Fusion \cite{yang2022vox}.
\subsubsection{Implemention Details}
Our approach is implemented using PyTorch. To enhance the efficiency of execution, our methodology incorporates a multi-process paradigm, segregating the tasks of tracking and mapping. These processes are concurrently executed on a pair of GTX3090 graphics cards. For the voxel framework, we have chosen a voxel dimension of 0.2, coupled with a step increment of 0.1, facilitating precise sampling within the voxels at the junctures of ray interactions.

\begin{table*}[!ht]
\renewcommand\arraystretch{1.2}
\resizebox{2.05\columnwidth}{!}{
\begin{tabular}{lllllllllllll}
\toprule 
\multicolumn{2}{l}{Methods}&Metric &Room0&Room1&Room2&Office0&Office1&Office2&Office3&Office4&Avg.\\  
\hline 
\multicolumn{2}{l}{\multirow{3}*{iMAP}}&\textbf{RMSE[cm]$\downarrow$}&70.05&4.53&2.20&2.32&1.74&4.87&5.84&2.62&1.83\\  
\multicolumn{2}{l}{~}&\textbf{mean[cm]$\downarrow$} &5.89&3.95&1.95&1.65&1.55&3.19&5.48&2.15&1.60\\
\multicolumn{2}{l}{~}&\textbf{mediam[cm]$\downarrow$} &44.78&3.35&1.73&1.35&1.37&2.35&47.56&1.86&1.30\\ 
\hline
\multicolumn{2}{l}{\multirow{3}*{NICE-SLAM}}&\textbf{RMSE[cm]$\downarrow$}&1.7&2.38&2.9&\cellcolor{green!20}\textbf{0.92}&\cellcolor{green!20}\textbf{0.87}&1.74&3.38&2.93&2.1\\  
\multicolumn{2}{l}{~}&\textbf{mean[cm]$\downarrow$} &1.47&2.0&1.56&\cellcolor{green!20}\textbf{0.8}&\cellcolor{green!20}\textbf{0.78}&1.48&2.2&2.17&1.54\\
\multicolumn{2}{l}{~}&\textbf{mediam[cm]$\downarrow$} &1.3&1.8&1.15&0.7&\cellcolor{green!20}\textbf{0.7}&1.29&1.55&1.69&1.27\\ 
\hline 
\multicolumn{2}{l}{\multirow{3}*{\textcolor{gray}{\textit{Vox-Fusion}}}}&\textcolor{gray}{\textit{RMSE[cm]$\downarrow$}}&\textcolor{gray}{\textit{0.27}}&\textcolor{gray}{\textit{1.33}}&\textcolor{gray}{\textit{0.47}}&\textcolor{gray}{\textit{0.70}}&\textcolor{gray}{\textit{1.11}}&\textcolor{gray}{\textit{0.46}}&\textcolor{gray}{\textit{0.26}}&\textcolor{gray}{\textit{0.58}}&\textcolor{gray}{\textit{N/A}}\\  
\multicolumn{2}{l}{~}&\textcolor{gray}{\textit{mean[cm]$\downarrow$}}&\textcolor{gray}{\textit{0.22}}&\textcolor{gray}{\textit{1.07}}&\textcolor{gray}{\textit{0.30}}&\textcolor{gray}{\textit{0.49}}&\textcolor{gray}{\textit{0.73}}&\textcolor{gray}{\textit{0.37}}&\textcolor{gray}{\textit{0.22}}&\textcolor{gray}{\textit{0.40}}&\textcolor{gray}{\textit{N/A}}\\
\multicolumn{2}{l}{~}&\textcolor{gray}{\textit{mediam[cm]$\downarrow$}}&\textcolor{gray}{\textit{0.20}}&\textcolor{gray}{\textit{0.78}}&\textcolor{gray}{\textit{0.27}}&\textcolor{gray}{\textit{0.29}}&\textcolor{gray}{\textit{0.45}}&\textcolor{gray}{\textit{0.31}}&\textcolor{gray}{\textit{0.19}}&\textcolor{gray}{\textit{0.28}}&\textcolor{gray}{\textit{N/A}}\\  \hline 
\multicolumn{2}{l}{\multirow{3}*{Vox-Fusion*}}&\textbf{RMSE[cm]$\downarrow$}&\cellcolor{green!20}\textbf{0.64}&1.13&\cellcolor{green!20}\textbf{0.87}&10.47&1.07&\cellcolor{green!20}\textbf{1.94}&1.14&\cellcolor{green!20}\textbf{1.08}&2.17\\  
\multicolumn{2}{l}{~}&\textbf{mean[cm]$\downarrow$} &\cellcolor{green!20}\textbf{0.57}&0.93&\cellcolor{green!20}\textbf{0.71}&6.43&0.92&1.84&1.06&\cellcolor{green!20}\textbf{0.98}&1.57\\
\multicolumn{2}{l}{~}&\textbf{mediam[cm]$\downarrow$} &0.53&0.79&\cellcolor{green!20}\textbf{0.63}&3.97&0.81&1.79&1.01&\cellcolor{green!20}\textbf{0.92}&1.24\\  \hline

\multicolumn{2}{l}{\multirow{3}*{Ours}}&\textbf{RMSE[cm]$\downarrow$}
&
0.83	&\cellcolor{green!20}\textbf{1.08}&	0.89&	1.48&	1.99&	1.99&	\cellcolor{green!20}\textbf{1.04}&	1.23&\cellcolor{green!20}\textbf{1.32}\\  
\multicolumn{2}{l}{~}&\textbf{mean[cm]$\downarrow$} &
0.70	&\cellcolor{green!20}\textbf{0.88}	&0.77	&0.92	&1.81&	\cellcolor{green!20}\textbf{1.75}&	\cellcolor{green!20}\textbf{0.97}&	1.08	&\cellcolor{green!20}\textbf{1.11}\\
\multicolumn{2}{l}{~}&\textbf{mediam[cm]$\downarrow$} &
0.61	 &\cellcolor{green!20}\textbf{0.73}	 &0.66	 &\cellcolor{green!20}\textbf{0.69} &	1.78	 &\cellcolor{green!20}\textbf{1.46}	 &\cellcolor{green!20}\textbf{0.94}	 &0.96	 &\cellcolor{green!20}\textbf{0.98} 
\\
\bottomrule 
\end{tabular}
}
\\
\caption{Tracking result on the Replica Dataset. The data of iMAP, Vox-Fusion are from \cite{yang2022vox}, Vox-Fusion* and NICE-SLAM are implemented by the open-source code. We conducted five tests for each scene and calculated the average values. Our method outperforms all existing approaches, with the best results prominently highlighted as the \sethlcolor{green!20}\hl{\textbf{first}}.}
\label{table1}
\end{table*}

\begin{table}[!ht]
\renewcommand\arraystretch{1.5}
\resizebox{1.0\columnwidth}{!}{
\begin{tabular}{lllllllll}
\toprule 
\multicolumn{2}{l}{Methods}&0000&0059&0106&0169&0181&0207&Avg.\\  
\hline 
\multicolumn{2}{l}{DI-Fusion}&62.99&128.00&18.50&75.80&87.88&100.19&78.89\\ 
\multicolumn{2}{l}{NICE-SLAM}&12.00&14.00&\cellcolor{green!20}\textbf{7.90}&10.90&13.40&6.20&10.70\\  
\multicolumn{2}{l}{Vox-Fusion}&16.55&24.18&8.41&27.28&23.30&9.41&18.52\\ 
\multicolumn{2}{l}{{Ours}}&\cellcolor{green!20}\textbf{12.71}&\cellcolor{green!20}\textbf{9.70}&8.50&\cellcolor{green!20}\textbf{8.92}&\cellcolor{green!20}\textbf{12.72}&\cellcolor{green!20}\textbf{5.61}&\cellcolor{green!20}\textbf{9.68}\\  
\bottomrule 
\end{tabular}
}
\\
\caption{Tracking result on the ScanNet Dataset. The data of DI-Fusion, NICE-SLAM, and Vox-Fusion are from \cite{sandstrom2023point}.}
\vspace{-0.5cm}
\label{table2}
\end{table}

\begin{table*}[!ht]
\renewcommand\arraystretch{1.25}
\resizebox{2.0\columnwidth}{!}{
\begin{tabular}{lllllllllllll}
\toprule 
\multicolumn{2}{l}{Methods}&Metric &Room0&Room1&Room2&Office0&Office1&Office2&Office3&Office4&Avg.\\  
\hline 
\multicolumn{2}{l}{\multirow{3}*{iMAP}}&\textbf{Acc.[cm]$\downarrow$}&3.58&3.69&4.68&5.87&3.71&4.81&4.27&4.83&4.43\\  
\multicolumn{2}{l}{~}&\textbf{Comp.[cm]$\downarrow$ }&5.06&4.87&5.51&6.11&5.26&5.65&5.45&6.59&5.56\\
\multicolumn{2}{l}{~}&\textbf{Comp.Ratio[\textless5cm\%]$\uparrow$} &83.91&83.45&75.53&77.71&79.64&77.22&77.34&77.63&79.06\\
\hline 
\multicolumn{2}{l}{\multirow{3}*{NICE-SLAM}}&\textbf{Acc.[cm]$\downarrow$}&3.53&3.60&3.03&5.56&3.35&4.71&3.84&3.35& 3.87&\\  
\multicolumn{2}{l}{~}&\textbf{Comp.[cm]$\downarrow$} &3.40&3.62&3.27&4.55&4.03&3.94&3.99&4.15& 3.87&\\
\multicolumn{2}{l}{~}&\textbf{Comp.Ratio[\textless5cm\%]$\uparrow$} &86.05&80.75&87.23&79.34&82.13&80.35&80.55&82.88& 82.41&\\ 
\hline 
\multicolumn{2}{l}{\multirow{3}*{Vox-Fusion}}&\textbf{Acc.[cm]$\downarrow$}&2.55&2.25&4.26&2.49&3.68&4.78&4.15&3.35& 3.44&\\  
\multicolumn{2}{l}{~}&\textbf{Comp.[cm]$\downarrow$} &2.82&2.36&2.67&\cellcolor{green!20}\textbf{1.64}&\cellcolor{green!20}\textbf{1.78}&3.02&3.11&3.36& 2.60&\\
\multicolumn{2}{l}{~}&\textbf{Comp.Ratio[\textless5cm\%]$\uparrow$} &90.97&92.72&89.88&\cellcolor{green!20}\textbf{95.40}&\cellcolor{green!20}\textbf{93.91}&\cellcolor{green!20}\textbf{89.06}&87.59& 86.12&90.71\\  \hline 
\multicolumn{2}{l}{\multirow{3}*{Ours}}&\textbf{Acc.[cm]$\downarrow$}&\cellcolor{green!20}\textbf{2.06}	&\cellcolor{green!20}\textbf{1.91}	&\cellcolor{green!20}\textbf{2.1}	&\cellcolor{green!20}\textbf{1.94}	&2.17	&\cellcolor{green!20}\textbf{2.2}	&\cellcolor{green!20}\textbf{2.21}	&\cellcolor{green!20}\textbf{2.08}	&\cellcolor{green!20}\textbf{2.08}&\\
 
\multicolumn{2}{l}{~}&\textbf{Comp.[cm]$\downarrow$} 

&\cellcolor{green!20}\textbf{2.64}	&\cellcolor{green!20}\textbf{2.33}	&\cellcolor{green!20}\textbf{2.21}	&1.75	&2.28	&\cellcolor{green!20}\textbf{2.83}	&\cellcolor{green!20}\textbf{2.86}	&\cellcolor{green!20}\textbf{3.1}	&\cellcolor{green!20}\textbf{2.5}&\\

\multicolumn{2}{l}{~}&\textbf{Comp.Ratio[\textless5cm\%]$\uparrow$} &\cellcolor{green!20}\textbf{92.26}&\cellcolor{green!20}\textbf{93.07}&\cellcolor{green!20}\textbf{92.57}&94.61&92.06&88.98&\cellcolor{green!20}\textbf{88.65}&\cellcolor{green!20}\textbf{88.06}&\cellcolor{green!20}\textbf{91.28}&\\
\bottomrule 
\end{tabular}
}
\\  
\caption{Reconstruction result on the Replica Dataset. The data of iMAP, NICE-SLAM, and Vox-Fusion are from \cite{yang2022vox}. Our system demonstrates the most outstanding performance.}
\label{table3}
\end{table*}

\begin{figure}[!t] 
\centering %
\vspace{-0.3cm}
\includegraphics[width=0.5\textwidth]{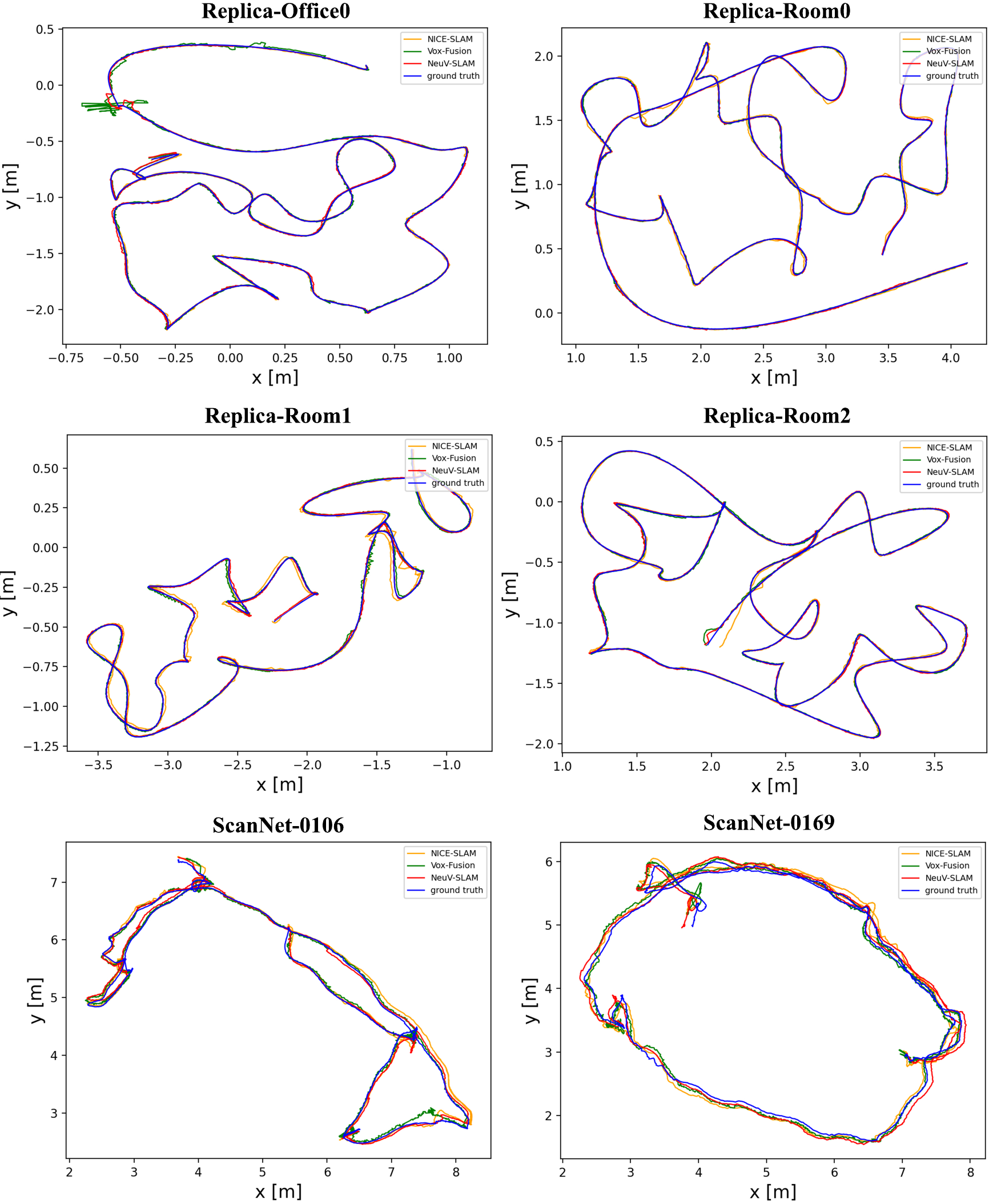} 
\caption{Qualitative tracking result on the Replica and ScanNet Datasets. 
We project the trajectory in three-dimensional space onto the x-y plane.} 
\vspace{-0.3cm}
\label{sdf} 
\end{figure}
\vspace*{-3mm}
\subsection{Tracking}
To assess the tracking accuracy of our system, we conducted a series of tracking experiments on both the Replica and ScanNet Datasets. The initial phase involved a qualitative comparison of the tracking trajectories produced by our system against the real trajectories recorded in the datasets. Specifically, for the Replica and ScanNet Datasets, the tracking trajectories were visually compared, as illustrated in Figure \ref{sdf}. The comparison revealed that our system demonstrates enhanced tracking performance over Vox-Fusion, particularly noticeable during the early stages of tracking and in situations involving sudden changes in the trajectory. This improvement can be largely attributed to the more sophisticated scene representation enabled by the use of multiresolution voxels. The incorporation of color information and SDF values directly into the voxel grid plays a pivotal role in this performance boost. This direct integration of descriptive features into the voxel structure allows for more rapid convergence of the system. As a result, our system is capable of quickly adapting to and accurately tracking changes in trajectory, even during the initial phases of network training or when faced with significant changes in the viewpoint. This leads to a more reliable and accurate tracking performance across various scenarios, underscoring the effectiveness of our approach in maintaining consistent trajectory tracking in complex environments.

We conducted a quantitative comparison of our system with NICE-SLAM \cite{zhu2022nice} and Vox-Fusion \cite{yang2022vox}. Table \ref{table1} presents the quantitative evaluation results on the Replica Dataset. These findings reveal that NeuV-SLAM demonstrates superior tracking performance compared to the current leading voxel-based methods. Our approach produces state-of-the-art results relative to these advanced voxel-based methods, further validating the efficacy of our method. As a real-world scene dataset, ScanNet poses greater challenges. Table \ref{table2} presents the tracking performance of NeuV-SLAM on the ScanNet Dataset, where it continues to demonstrate superior performance, outperforming competing methods.
\begin{figure}[!t] 
\centering %
\includegraphics[width=0.49\textwidth]{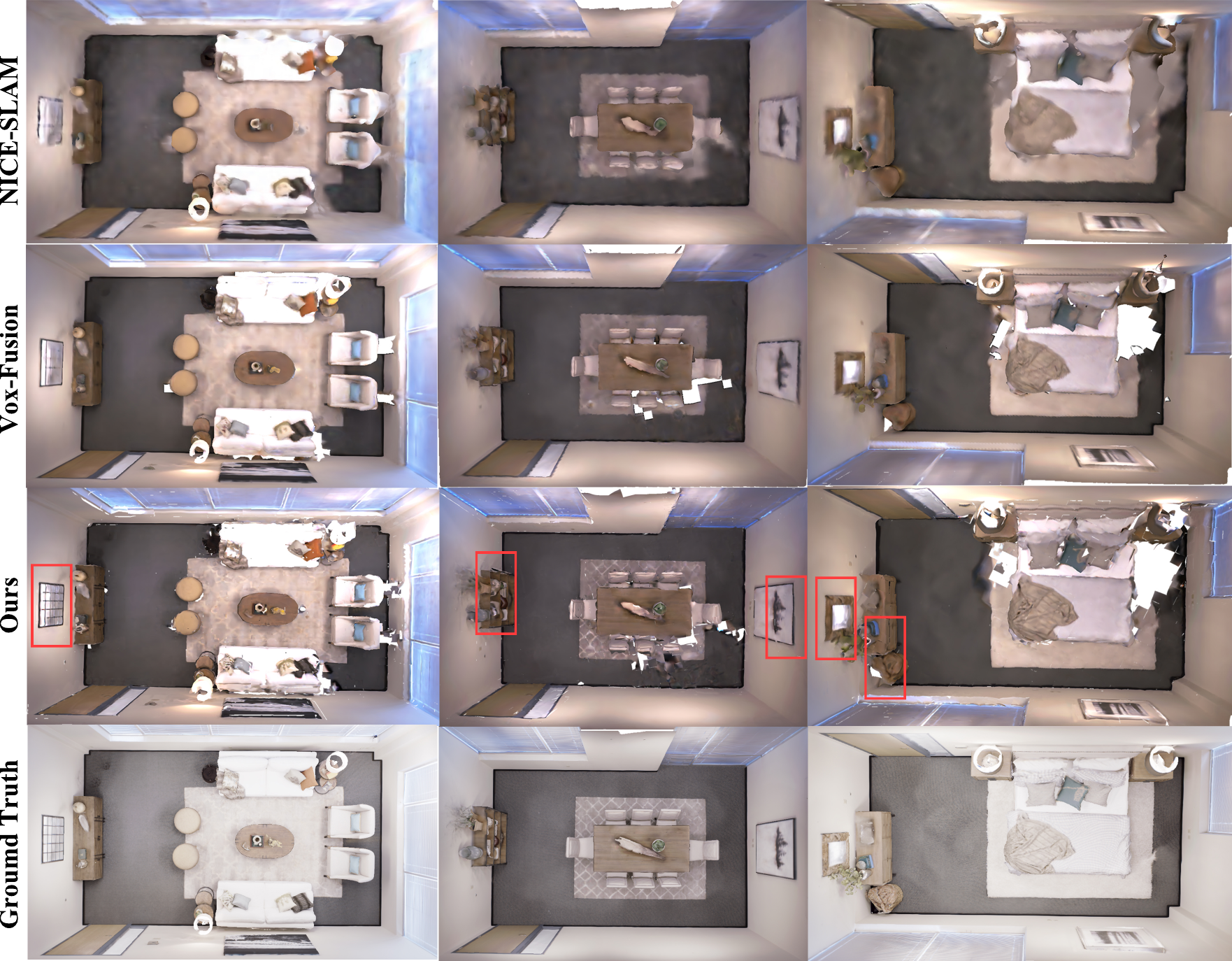} 
\caption{Qualitative reconstruction results on the Replica Dataset. From top to bottom, we show the results of scene reconstruction of different
methods (NICE-SLAM, Vox-Fusion, our method, and ground truth). 
To better visualize the differences in reconstruction between our method and other approaches, we employ red boxes to accentuate areas where our reconstruction method shows noticeable enhancements compared to other methods.} 
\vspace{-0.2cm}
\label{reco} 
\end{figure}
\vspace*{-3mm}
\subsection{Mapping}
We performed a qualitative comparison of our system with NICE-SLAM  \cite{zhu2022nice} and Vox-Fusion \cite{yang2022vox}. The results of this comparison focused on the Replica Dataset, are detailed in Figure \ref{reco}. A key aspect of NICE-SLAM is its initial assumption of a complete surface fill across the entire space. Consequently, this method generates surfaces even in areas that have not been observed. When the gaps are small, the surfaces created in this manner closely resemble the actual surfaces. However, in cases where the gaps are significant, there is a substantial deviation between the generated and the real surfaces. Vox-Fusion, on the other hand, utilizes a single-resolution grid, which often struggles to model surfaces with smaller structures. Our system, similar to Vox-Fusion, models surfaces only within visible sparse voxels. This approach allows us to generate realistic and reasonable hole-filling effects. Additionally, we employ dense voxels to capture detailed surface textures, setting our method apart from the limitations observed in single-resolution grid approaches like Vox-Fusion. Our refined methodology boosts the system's capability to precisely mimic and monitor complex surfaces, especially beneficial in settings characterized by detailed surface intricacies.
\begin{figure*}[!t] 
\centering %
\includegraphics[width=1.0\textwidth]{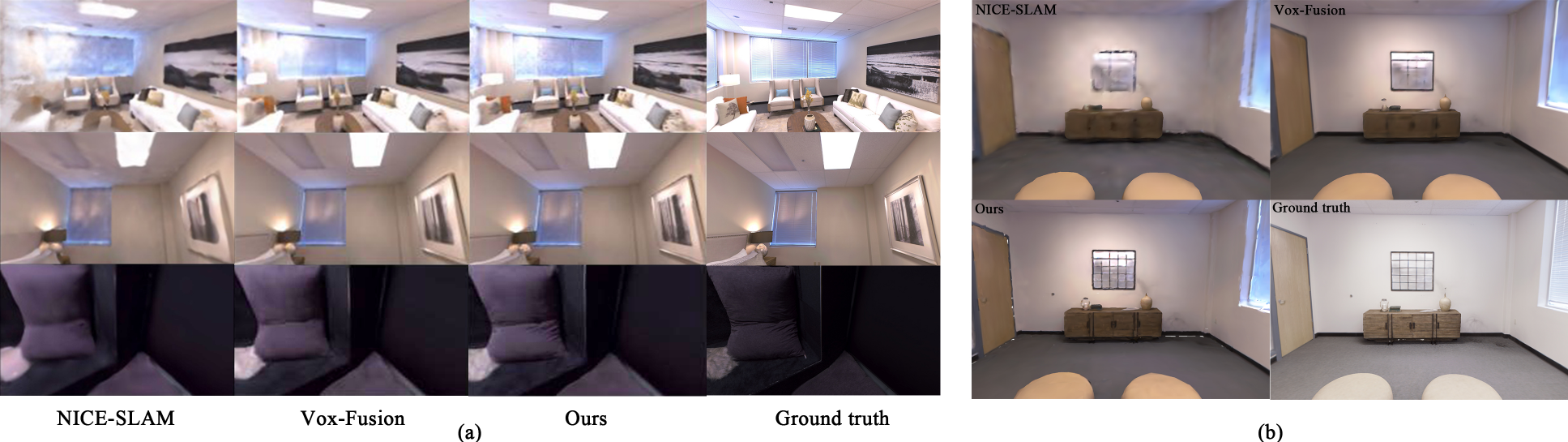} 
\caption{(a) Qualitative rendering result in Replica Dataset. Our method exhibits superior rendering results in some details. (b) Qualitative rendering performance in scene mesh. The windows and cabinets in the mesh have better rendering details.} 
\label{render1} 
\end{figure*}
\par
Table \ref{table3} quantitatively illustrates a comparative analysis of the existing iMAP, Vox-Fusion, NICE-SLAM, and our proposed method in the context of scene reconstruction on the Replica Dataset. The data for Vox-Fusion and NICE-SLAM are sourced from Vox-Fusion. Owing to the implementation of multiresolution voxel representation and the application of SDF activation mechanisms, our method demonstrates a significant advantage in the reconstruction of micro-level details. 
\vspace*{-7mm}
\subsection{Rendering}
% Figure \ref{render1} demonstrates the viewpoint rendering performance of our method, particularly highlighting the capacity of NeuV-SLAM to produce photo-realistic quality rendered images. Compared to Vox-Fusion and NICE-SLAM, our approach exhibits a significant superiority in rendering fine details in images.

In Figure \ref{render1}, the rendering capabilities of our NeuV-SLAM system are illustrated, demonstrating its proficiency in generating images with photo-realistic quality. Our approach stands out by rendering fine details more effectively than Vox-Fusion and NICE-SLAM, owing to its sophisticated multi-resolution voxel representation and neural processing techniques. This allows for a richer, more detailed visual output, capturing the nuances of the scene with greater fidelity.

Additionally, Table \ref{render} presents a quantitative analysis of rendering accuracy on the Replica Dataset. Compared to existing voxel-based advanced techniques, our method achieves state-of-the-art performance on the Replica Dataset.
\begin{table*}[!ht]
\renewcommand\arraystretch{1.2}
\resizebox{2.0\columnwidth}{!}{
\begin{tabular}{lllllllllllll}
\toprule 
\multicolumn{2}{l}{Methods}&Metric &Room0&Room1&Room2&Office0&Office1&Office2&Office3&Office4&Avg.\\  
\hline 
\multicolumn{2}{l}{\multirow{3}*{NICE-SLAM}}&\textbf{PSNR[dB]$\uparrow$}&22.12&22.47&24.52&29.07&30.34&19.66&22.23&24.94&24.42&\\  
\multicolumn{2}{l}{~}&\textbf{SSIM$\uparrow$} &0.689&0.757&0.814&0.874&0.886&0.797&0.801&0.856& 0.809&\\
\multicolumn{2}{l}{~}&\textbf{LPIPS$\downarrow$} &0.330&0.271&0.208&0.229&0.181&0.235&0.209&0.198& 0.233&\\ 
\hline 
\multicolumn{2}{l}{\multirow{3}*{Vox-Fusion}}&\textbf{PSNR[dB]$\uparrow$}&22.39&22.36&23.92&27.79&29.83&20.33&23.47&25.21& 24.41&\\  
\multicolumn{2}{l}{~}&\textbf{SSIM$\uparrow$}&0.683&0.751&0.798&0.857&0.876&0.794&0.803&0.847&0.801&\\
\multicolumn{2}{l}{~}&\textbf{LPIPS$\downarrow$} &0.303&0.269&0.234&0.241&0.184&0.243&0.213&0.199& 0.236&\\  \hline

\multicolumn{2}{l}{\multirow{3}*{Ours}}&\textbf{PSNR[dB]$\uparrow$}
&\cellcolor{green!20}\textbf{27.17}&\cellcolor{green!20}\textbf{29.24}&\cellcolor{green!20}\textbf{29.13}&\cellcolor{green!20}\textbf{32.98}&\cellcolor{green!20}\textbf{33.41}&\cellcolor{green!20}\textbf{27.18}&\cellcolor{green!20}\textbf{27.69}&\cellcolor{green!20}\textbf{30.71}&\cellcolor{green!20}\textbf{29.69}&\\  
\multicolumn{2}{l}{~}&\textbf{SSIM$\uparrow$}
&\cellcolor{green!20}\textbf{0.772}& \cellcolor{green!20}\textbf{0.813}&\cellcolor{green!20}\textbf{ 0.882} &\cellcolor{green!20}\textbf{0.883} &\cellcolor{green!20}\textbf{0.896 }&\cellcolor{green!20}\textbf{0.860}& \cellcolor{green!20}\textbf{0.842 }&\cellcolor{green!20}\textbf{0.879}&\cellcolor{green!20}\textbf{0.853} \\
\multicolumn{2}{l}{~}&\textbf{LPIPS$\downarrow$} 
&\cellcolor{green!20}\textbf{0.223}& \cellcolor{green!20}\textbf{0.197} &\cellcolor{green!20}\textbf{0.211}& \cellcolor{green!20}\textbf{0.208}& \cellcolor{green!20}\textbf{0.147} &\cellcolor{green!20}\textbf{0.214}& \cellcolor{green!20}\textbf{0.189}& \cellcolor{green!20}\textbf{0.157}&\cellcolor{green!20}\textbf{0.193 } \\
\bottomrule 
\end{tabular}
}
\\  
\caption{Rendering result on the Replica Dataset. The data of NICE-SLAM and Vox-Fusion are from \cite{zhu2023nicer}.}
\vspace{-0.3cm}
\label{render}
\end{table*}
\vspace*{-3mm}
\subsection{Performance Analysis}
\subsubsection{Convergence Speed}
To accurately assess the convergence efficiency of the system developed in this study, two distinct evaluation strategies were employed. Firstly, real-time scene rendering was implemented during the training phase of the neural network. This was to compare rendering quality with an equivalent amount of training data and the same batch size, meaning rendering was conducted at every keyframe interval. This step primarily served for the quantitative analysis of the network's convergence speed. Specifically, our method was compared in detail with Vox-Fusion using the Replica Dataset, and the related data results have been presented in the subsequent table \ref{table5}. These results indicate that with the same amount of training data, our method stands out for its exceptional image rendering quality, highlighting the rapid convergence capability of our approach.
\begin{figure}[!t] 
\centering %
\includegraphics[width=0.5\textwidth]{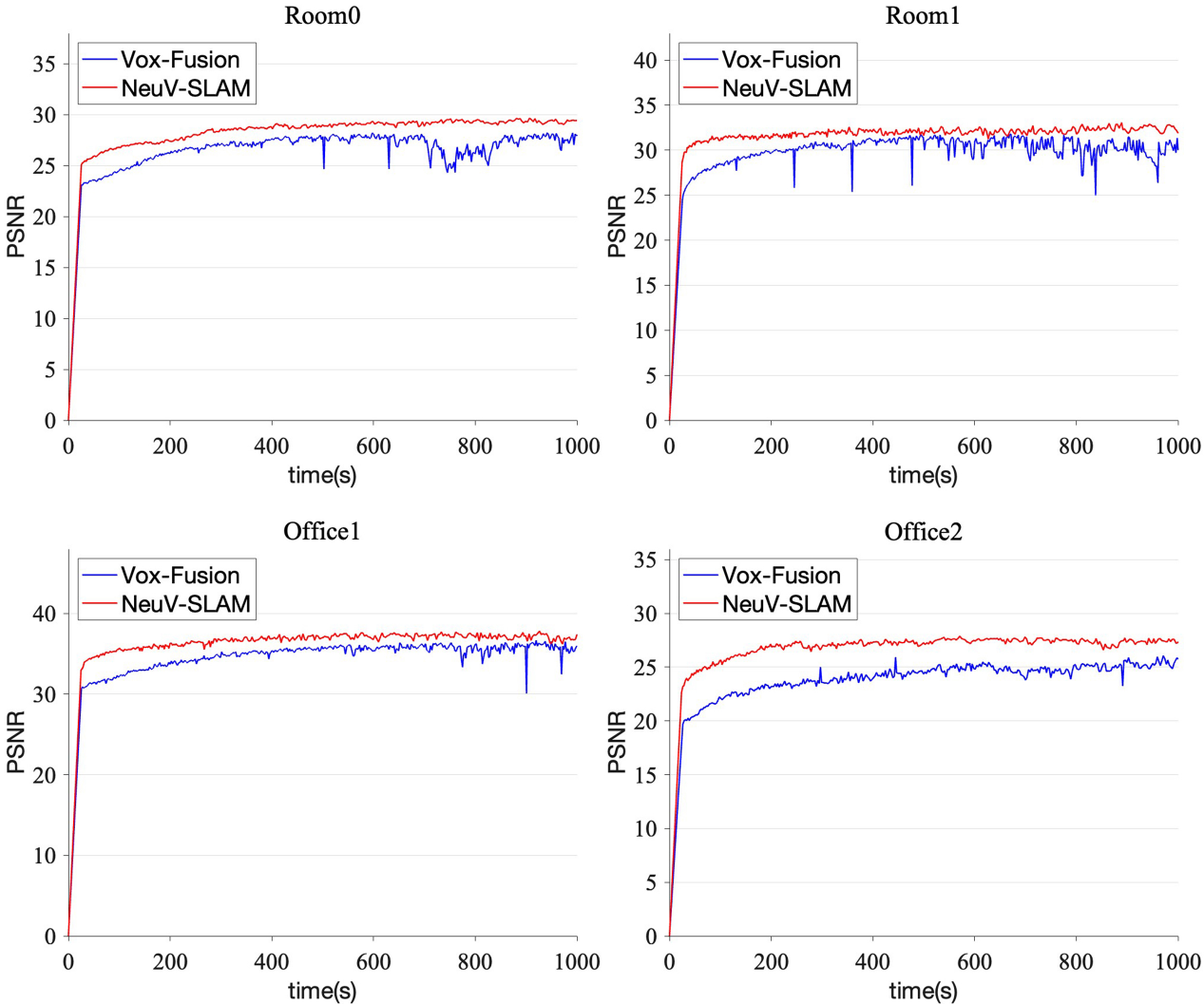} 
\caption{
Evaluation of the convergence speed between Vox-Fusion and our method. The horizontal axis is the system running time, and the vertical axis is the rendering PSNR value for the same viewpoint.} 
\vspace{-0.35cm}
\label{speed} 
\end{figure}
\par
We also implemented a second evaluation technique, which entailed periodically rendering images from a consistent viewpoint throughout the training period. This approach aimed to monitor how the quality of the images evolved in relation to the length of the training process. The findings from this experiment revealed that our method not only reached convergence more swiftly than the Vox-Fusion technology but also produced images of superior rendering quality upon completion of training, as depicted in Figure \ref{speed}. This demonstrates the efficiency and effectiveness of our approach in achieving high-quality results in a shorter timeframe.
\begin{table}[t]
\centering
\renewcommand\arraystretch{1.3}
\resizebox{1.0\columnwidth}{!}{
\begin{tabular}{llllllllll}
\toprule 
Methods&Ro0&Ro1&Ro2&Of0&Of1&Of2&Of3&Of4&Avg.\\  
\hline  
Vox-Fusion&26.06&28.29&28.61&29.93&32.28&26.65&26.32&29.37&28.43\\  
\hline 
Ours&\cellcolor{green!20}\textbf{26.37}&\cellcolor{green!20}\textbf{28.31}&\cellcolor{green!20}\textbf{28.43}&\cellcolor{green!20}\textbf{32.07}&\cellcolor{green!20}\textbf{32.81}&\cellcolor{green!20}\textbf{26.88}&\cellcolor{green!20}\textbf{26.57}&\cellcolor{green!20}\textbf{29.57}&\cellcolor{green!20}\textbf{28.88}\\  
\bottomrule 
\\
\end{tabular}
}
\caption{
Rendering results under the same training volume on the Replica Dataset.  }
\vspace{-0.8cm}
\label{table5}
\end{table}
\subsubsection{Time and Memory Efficiency}
To provide an in-depth analysis of the impact of multiresolution voxel-based sampling and rendering techniques on system runtime, this study conducted a systematic performance evaluation on the Replica Dataset. Key experimental procedures included testing the time required for the tracking and mapping stages during a single iteration. The detailed results are presented in Table \ref{table6}. The findings indicate that although our method is slightly slower than Vox-Fusion during the sampling phase—due to the need to independently sample sparse and dense voxels before merging, a more time-consuming process compared to Vox-Fusion's single sampling—our technique's efficiency in scene convergence allows us to significantly reduce the overall time consumption by decreasing the number of iterations while maintaining performance comparable to Vox-Fusion.
\begin{table}[!ht]
\centering
\setlength{\tabcolsep}{10pt}
\renewcommand\arraystretch{1.8}
% \renewcommand\arraystretch{0.8}
% \resizebox{0.5\columnwidth}{!}{
\begin{tabular}{ccc}
\toprule 
Methods&Tracking&Mapping\\  
\hline 
\textcolor{gray}{\textit{Vox-Fusion}}&\textcolor{gray}{\textit{12ms}}&\textcolor{gray}{\textit{55ms}}\\  
\hline 
Vox-Fusion*&25ms&22ms\\  
\hline 
ours&28ms&24ms\\  
\bottomrule 
\\
\end{tabular}
% }
% \\
\caption{The time of tracking and mapping during a single iteration. The data of Vox-Fusion are from \cite{yang2022vox}. The data of Vox-Fusion* are implemented by the open-source code. }
\vspace{-0.3cm}
\label{table6}
\end{table}
% \vspace*{-3mm}
\par
Additionally, the study evaluated the memory consumption of the implicit scene decoder and voxel embeddings in the system, comparing them with Vox-Fusion and NICE-SLAM in the Replica office-0 scene. As shown in Table \ref{table7}, our NeuV-SLAM method consumes more memory for voxel embeddings than Vox-Fusion, due to our choice to explicitly store SDF values in voxels, thereby increasing memory usage. In contrast, due to its four-layer voxel structure, NICE-SLAM exhibits the most significant memory consumption. \par
\begin{table}[!ht]
\centering
\setlength{\tabcolsep}{10pt}
\renewcommand\arraystretch{1.6}
% \renewcommand\arraystretch{0.8}
% \resizebox{0.5\columnwidth}{!}{
\begin{tabular}{ccc}
\toprule 
Methods&Decoder&Embedding\\  
\hline 
NICE-SLAM&0.22MB&238.88MB\\ 
\hline 
\textcolor{gray}{\textit{Vox-Fusion}}&\textcolor{gray}{\textit{1.04MB}}&\textcolor{gray}{\textit{0.149MB}}\\  
\hline 
Vox-Fusion*&0.43MB&1.22MB\\  
\hline 
ours&0.15MB&1.29MB\\  
\bottomrule 
\\
\end{tabular}
% }
% \\
\caption{ The memory consumption on the Replica Dataset. The data of NICE-SLAM and Vox-Fusion are from \cite{yang2022vox}. The data of Vox-Fusion* are implemented by the open-source code.}
\label{table7}
\vspace{-0.2cm}
\end{table}
% \vspace*{-3mm}
It is noteworthy that the memory footprint of our decoder part is less. This is because our decoder is only used for decoding color features, resulting in a more streamlined network structure—our network consists of three layers, while Vox-Fusion employs a six-layer structure. Consequently, our approach not only significantly reduces memory usage but also achieves higher reconstruction accuracy.
\begin{figure}[!t] 
\centering %
\vspace{-0.15cm}
\includegraphics[width=0.46\textwidth]{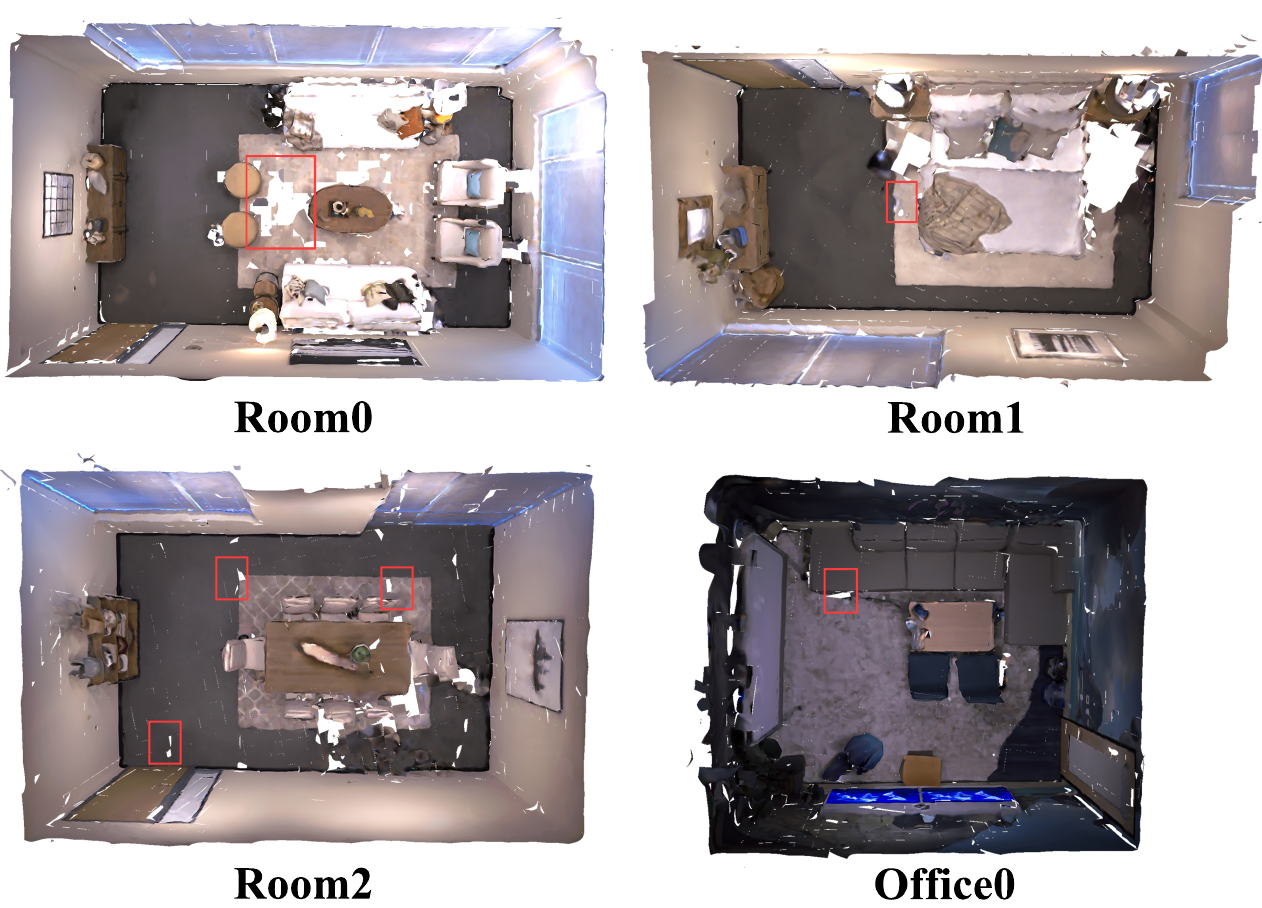} 
\caption{Qualitative reconstruction result without SDF activation.} 
\vspace{-0.4cm}
\label{recon} 
\end{figure}
\vspace*{-3mm}
\subsection{Ablation Study}
\subsubsection{Effectiveness of SDF Activation}
To verify the effectiveness of the SDF activation strategy, this study conducted ablation experiments on the SDF activation strategy. By removing the SDF activation function, the experiments aimed to evaluate its impact on the system's tracking and mapping performance. The preliminary research focused on a qualitative analysis of the SDF activation function. Figure \ref{recon} presents a comparison of rendering quality with and without the SDF activation function. The results clearly indicate that without applying SDF activation strategies, the system's ability to fit the scene is limited, resulting in some voids or gaps. This activation function, through the introduction of nonlinear estimation, improves the system's ability to fit multiresolution voxel scenes, thereby achieving better reconstruction results.\par
Furthermore, Table \ref{table11} provides quantitative data on system performance following the removal of the SDF activation strategy. By comparing the data in Table \ref{table1} and \ref{table3}, the notable benefits of the SDF activation strategy in camera pose tracking and scene reconstruction can be observed. The activation function, by improving the capture of detailed information, accelerates the overall scene's convergence process and positively impacts camera pose tracking.

\begin{table*}[!ht]
\renewcommand\arraystretch{1.2}
\resizebox{2.0\columnwidth}{!}{
\begin{tabular}{lllllllllllll}
\toprule 
\multicolumn{2}{l}{}&Metric &Room0&Room1&Room2&Office0&Office1&Office2&Office3&Office4&Avg.\\  
\hline 
\multicolumn{2}{l}{\multirow{3}*{Tracking}}&\textbf{RMSE[cm]$\downarrow$}&0.72&2.25&1.1&1.16&0.87&1.64&0.84&5.72&1.78\\  
\multicolumn{2}{l}{~}&\textbf{mean[cm]$\downarrow$} &0.58&1.32&0.87&0.83&2.39&1.40&1.48&5.04&1.74\\
\multicolumn{2}{l}{~}&\textbf{mediam[cm]$\downarrow$} &0.47&0.75&0.73&0.67&2.24&1.19&1.29&4.95&1.54\\ 
\hline 
\multicolumn{2}{l}{\multirow{3}*{Reconstruction}}&\textbf{Acc.[cm]$\downarrow$}&2.73& 2.99&2.79&2.36&2.94&2.61&2.72&2.52&2.73 \\  
\multicolumn{2}{l}{~}&\textbf{Comp.[cm]$\downarrow$} &2.74&2.36&2.32&1.76&2.47&2.90&2.68&2.89&2.51 \\
\multicolumn{2}{l}{~}&\textbf{Comp.Ratio[\textless5cm\%]$\uparrow$} &90.25&91.81&90.52&93.90&90.56&87.65&89.64&87.16&90.31 \\   
\bottomrule 
\end{tabular}
}
\\
\caption{Tracking ann Reconstruction result on the Replica Dataset of NeuV-SLAM without SDF activation.}
\label{table11}
\end{table*}
\subsubsection{Effectiveness of  Multiresolution Voxel}
To validate the efficacy of multiresolution voxels in enhancing performance and reducing memory consumption, we conducted an ablation study by employing single-resolution voxels to compare the system's tracking and mapping capabilities. Table \ref{table8},\ref{table9} quantitatively demonstrates the performance in tracking and mapping at single resolutions of 0.2 and 0.1. As indicated by the table, the system's performance at a resolution of 0.1 is notably superior to that at 0.2, illustrating the direct impact of resolution size on tracking and mapping. Comparing the performance at a resolution of 0.1 with multiresolution performance (Table \ref{table1},\ref{table3}), we observe a close similarity in performance, affirming the reliability of the multiresolution strategy in enhancing system performance. Furthermore, We measured the memory consumption of the 0.1-resolution voxel embedding for the Replica room0 scene, which amounted to 1.52 MB. This represents higher memory consumption compared to Table \ref{table7}. Multiresolution voxels, in contrast to dense multiresolution voxels, consume less memory, enabling the system to operate in large-scale scenarios. \par
\begin{table*}[!ht]
\renewcommand\arraystretch{1.3}
\resizebox{2.0\columnwidth}{!}{
\begin{tabular}{llllllllllll}
\toprule 
\multicolumn{2}{l}{}&Metric &Room0&Room1&Room2&Office0&Office1&Office2&Office3&Office4&Avg.\\  
\hline 
\multicolumn{2}{l}{\multirow{3}*{0.1 resolution}}&\textbf{RMSE[cm]$\downarrow$}
&\cellcolor{green!20}\textbf{0.66}	&\cellcolor{green!20}\textbf{0.83}&	\cellcolor{green!20}\textbf{1.15}	&\cellcolor{green!20}\textbf{1.18}	&\cellcolor{green!20}\textbf{1.98	}&\cellcolor{green!20}\textbf{1.76	}&\cellcolor{green!20}\textbf{0.95}	&\cellcolor{green!20}\textbf{1.24 }&\cellcolor{green!20}\textbf{1.20}
\\  
\multicolumn{2}{l}{~}&\textbf{mean[cm]$\downarrow$} 
&\cellcolor{green!20}\textbf{0.58}&	\cellcolor{green!20}\textbf{0.71}	&\cellcolor{green!20}\textbf{0.91}	&\cellcolor{green!20}\textbf{0.89}	&\cellcolor{green!20}\textbf{1.83}	&\cellcolor{green!20}\textbf{1.57}	&\cellcolor{green!20}\textbf{0.88}&	\cellcolor{green!20}\textbf{1.16}	&\cellcolor{green!20}\textbf{1.07}\\
\multicolumn{2}{l}{~}&\textbf{mediam[cm]$\downarrow$} 
&\cellcolor{green!20}\textbf{0.52}	&\cellcolor{green!20}\textbf{0.63}	&\cellcolor{green!20}\textbf{0.83}	&\cellcolor{green!20}\textbf{0.78	}&\cellcolor{green!20}\textbf{1.78}	&\cellcolor{green!20}\textbf{1.33}	&\cellcolor{green!20}\textbf{0.86}	&\cellcolor{green!20}\textbf{1.12}	&\cellcolor{green!20}\textbf{0.95}\\ 
\hline 
\multicolumn{2}{l}{\multirow{3}*{0.2 resolution}}&\textbf{RMSE[cm]$\downarrow$}
&1.16	&1.24&	1.32	&1.39	&2.84	&3.23	&1.32	&1.68	&1.76\\
\multicolumn{2}{l}{~}&\textbf{mean[cm]$\downarrow$} 
&1.01	&1.12	&1.14	&1.07&	2.62	&2.85	&1.26	&1.52&	1.69\\
\multicolumn{2}{l}{~}&\textbf{mediam[cm]$\downarrow$} 
&0.92&	1.06	&1.02	&0.91	&2.61	&2.25	&1.27	&1.41	&1.43\\

\bottomrule 
\end{tabular}
}
\\
\caption{Tracking result on the Replica Dataset in different voxel resolution.}
\label{table8}
\end{table*}

\begin{table*}[!ht]
\renewcommand\arraystretch{1.3}
\resizebox{2.0\columnwidth}{!}{
\begin{tabular}{lllllllllllll}
\toprule 
\multicolumn{2}{l}{}&Metric &Room0&Room1&Room2&Office0&Office1&Office2&Office3&Office4&Avg.\\  
\hline 
\multicolumn{2}{l}{\multirow{3}*{0.1 resolution}}&\textbf{Acc.[cm]$\downarrow$}&
\cellcolor{green!20}\textbf{1.56}&
\cellcolor{green!20}\textbf{1.33}&
\cellcolor{green!20}\textbf{1.55}&
\cellcolor{green!20}\textbf{1.90}&
\cellcolor{green!20}\textbf{1.22}&
\cellcolor{green!20}\textbf{1.67}&
\cellcolor{green!20}\textbf{1.79}&
\cellcolor{green!20}\textbf{1.74}&
\cellcolor{green!20}\textbf{1.59}&\\  
\multicolumn{2}{l}{~}&\textbf{Comp.[cm]$\downarrow$}&
\cellcolor{green!20}\textbf{3.05}&
\cellcolor{green!20}\textbf{2.50}&
\cellcolor{green!20}\textbf{2.40}&
\cellcolor{green!20}\textbf{1.71}&
\cellcolor{green!20}\textbf{1.93}&
\cellcolor{green!20}\textbf{3.19}&
\cellcolor{green!20}\textbf{3.11}&
3.51&
\cellcolor{green!20}\textbf{2.67}&\\
\multicolumn{2}{l}{~}&\textbf{Comp.Ratio[\textless5cm\%]$\uparrow$}&
\cellcolor{green!20}\textbf{90.72}&
\cellcolor{green!20}\textbf{91.82}&
\cellcolor{green!20}\textbf{91.51}&
\cellcolor{green!20}\textbf{94.88}&
\cellcolor{green!20}\textbf{93.30}&
\cellcolor{green!20}\textbf{88.42}&
\cellcolor{green!20}\textbf{87.85}&
\cellcolor{green!20}\textbf{86.99}&
\cellcolor{green!20}\textbf{90.68}&
\\ 
\hline 
\multicolumn{2}{l}{\multirow{3}*{0.2 resolution}}&\textbf{Acc.[cm]$\downarrow$}&2.21&1.96&2.11&2.08&2.58&2.46&2.26&2.21& 2.23&\\  
\multicolumn{2}{l}{~}&\textbf{Comp.[cm]$\downarrow$} &3.08&2.56&2.50&2.01&2.58&3.69&3.23&\cellcolor{green!20}\textbf{3.35}& 2.87&\\
\multicolumn{2}{l}{~}&\textbf{Comp.Ratio[\textless5cm\%]$\uparrow$}&
89.49&
91.40&
89.77&
92.41&
89.78&
84.72&
85.66&
86.52&
88.71&\\  
\bottomrule 
\end{tabular}
}
\\  
\caption{Reconstruction result on the Replica Dataset in different voxel resolution.}
\vspace{-0.3cm}
\label{table9}
\end{table*}

\subsubsection{Different  Threshold of edge detection}
\begin{table}[t]
\renewcommand\arraystretch{1.4}
\vspace{-0.1cm}
\resizebox{1\columnwidth}{!}{
\begin{tabular}{llccccccccc}
\toprule 
\multicolumn{2}{l}{} &30:60&30:90&50:100&50:150&70:140&70:210&90:180&90:270\\ 
\hline 
\multicolumn{2}{l}{\multirow{3}*{Tracking}}&
\cellcolor{green!20}\textbf{0.69}&0.70&0.74&0.78&0.73&0.79&0.87&0.87&\\  
\multicolumn{2}{l}{~}&\cellcolor{green!20}\textbf{0.58}&0.60&0.60&0.62&0.61&0.64&0.72&0.74&\\
\multicolumn{2}{l}{~}&\cellcolor{green!20}\textbf{0.49}&0.54&0.52&0.53&0.53&0.54&0.65&0.66&\\ 
\hline 
\multicolumn{2}{l}{\multirow{3}*{Reconstruction}}&
2.06&
2.17&
2.13&
2.02&
2.07&
\cellcolor{green!20}\textbf{2.05}&
2.13&
2.14&\\  
\multicolumn{2}{l}{~}&
\cellcolor{green!20}\textbf{2.64}&
2.66&
2.69&
2.66&
2.67&
2.67&
2.68&
2.69&\\
\multicolumn{2}{l}{~}&
\cellcolor{green!20}\textbf{92.26}&
92.025&
92.28&
92.02&
92.06&
92.18&
92.08&
91.88&\\  
\bottomrule 
\end{tabular}
}
\\
\caption{Tracking ann Reconstruction result on the Replica Dataset of NeuV-SLAM with different  threshold}
\vspace{-0.5cm}
\label{table10}
\end{table}
We generate dense voxels by detecting two-dimensional information and extracting edges, employing the classic Canny algorithm \cite{canny1983finding} for edge extraction. The Canny algorithm necessitates two thresholds to jointly fine-tune edge detection. Excessively high thresholds result in a paucity of detected edges, whereas too-low thresholds lead to the appearance of false edges. We explored the influence of varying threshold values on the generation of dense voxels and their subsequent impact on system performance. Typically, in the Canny algorithm, the ratio of high to low threshold is set between 3:1 and 2:1. For our experiments, we started with 30 and increased by increments of 20 in the Replica Room0 scenario to compare performance, as shown in Table \ref{table10}. The results indicate that the optimal performance was achieved with thresholds set at 30:60. The superior performance at this lower threshold of 30:60 demonstrates that the generation of more dense voxels positively affects system performance.
% \vspace*{-1mm}
\section{conclusion}
In conclusion, we have explored the inherent limitations of traditional SLAM methods in accurately capturing detailed information and the difficulties associated with efficiently incrementally expanding scenes in NeRF-based SLAM systems. To address these challenges, we introduced NeuV-SLAM, a novel dense SLAM system based on neural multiresolution voxels. The key innovations of this research encompass the development of an efficient hash-based multiresolution voxel generation and management structure, hashMV, which facilitates rapid dynamic scene expansion while maintaining a compact memory footprint. Additionally, we propose a novel implicit representation method, anchoring neural features and SDF values directly within voxels, employing SDF activation strategy for efficient scene convergence and enhanced scene representation. Our method was rigorously evaluated on RGBD datasets, showcasing its competitive performance in terms of convergence speed, reconstruction quality, localization accuracy, and rendering capabilities. The progress made in enhancing SLAM technologies not only addresses the current limitations of SLAM methodologies but also opens up avenues for broader applications in fields such as robotics, autonomous navigation, and augmented reality. 
\bibliographystyle{IEEEtran}
\bibliography{reference}
\vfill

\end{document}